\documentclass{article}

\usepackage{liquid_style}
\usepackage{liquid_color}
\usepackage{booktabs}
\usepackage{array}
\usepackage{tabularx}
\usepackage{pgfplots}
\pgfplotsset{compat=1.18}
\usetikzlibrary{arrows.meta, positioning, shapes.geometric, fit, backgrounds}

\usepackage{listings}
\usepackage{placeins}
\newcommand{\RepoRowURL}[2]{%
  \raisebox{-0.15\height}{\includegraphics[height=1.30em]{#1}}%
  & \href{#2}{\texttt{#2}}\\[-0.05em]
}

\title{In-Place Tokenizer Expansion for Pre-trained LLMs}
\author{%
  Jimmy T.H. Smith, Tarek Dakhran, Alberto Cabrera, Simon S. Lee, Paul Pak,\\
  Aditya Tadimeti, Tim Seyde, Maxime Labonne, Alexander Amini, Mathias Lechner\\[0.35em]
  Liquid AI}
\date{}

\usepackage{hyperref}

\newcommand{\reportdate}{\today}
\newcommand{\correspondingemail}{\href{mailto:jimmy@liquid.ai}{\texttt{jimmy@liquid.ai}}, \href{mailto:mathias@liquid.ai}{\texttt{mathias@liquid.ai}}}

\let\origfootrule\footrule
\renewcommand{\footrule}{\iffootnote{}{\origfootrule}}

\begin{document}
\maketitle
\renewcommand{\thefootnote}{\fnsymbol{footnote}}%
\renewcommand{\thefootnote}{\arabic{footnote}}\setcounter{footnote}{0}

\begin{center}
\setlength{\tabcolsep}{4pt}
\renewcommand{\arraystretch}{1.05}
\begin{tabular}{@{}>{\centering\arraybackslash}m{1.6em}@{\hspace{0.35em}}l@{}}
\RepoRowURL{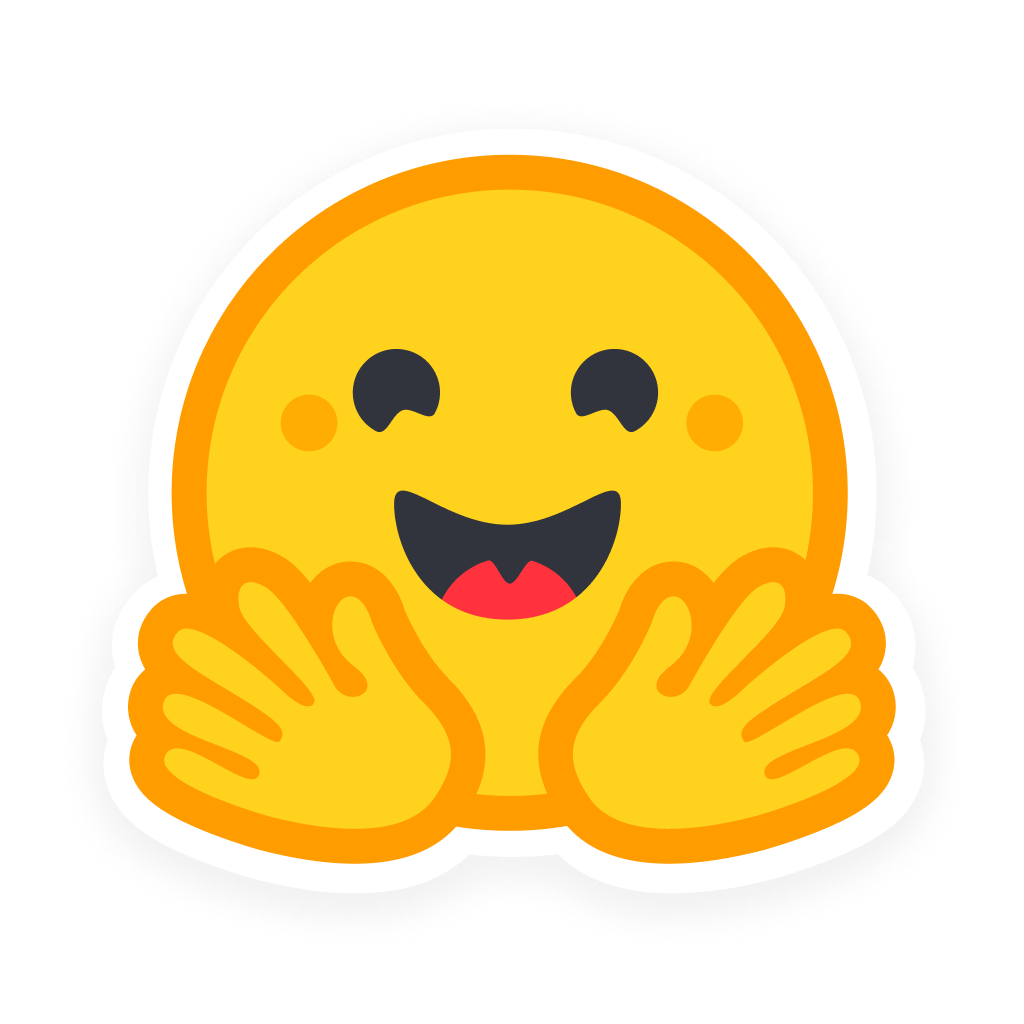}{https://huggingface.co/LiquidAI}
\end{tabular}
\end{center}
\vspace{1.2\baselineskip}

\begin{abstractbox}
A tokenizer fixed at the start of pre-training allocates vocabulary in proportion to the pre-training corpus, reflecting the deployment priorities at that time. When those priorities shift, languages added later are split into many more tokens per word, which can raise latency, compute, and energy consumption for users of those languages. Cloud models can afford a broad vocabulary because the embedding and LM-head matrices are a small fraction of their parameters. On a compact model those matrices are a material share of per-token decode bandwidth, so on-device models ship small vocabularies and accept fragmentation outside a fixed language set. We present \emph{tokenizer expansion}, an in-place recipe for upgrading a pre-trained model's tokenizer when the model producer controls its design. We continue the existing tokenizer's BPE merges on a multilingual corpus, so most source tokens carry over unchanged as single tokens and every new token has an exact decomposition into source tokens. We copy the carried-over embedding rows unchanged and initialize new rows as the mean of their source sub-token embeddings. A two-stage adaptation, embedding-only training then full-model continued pre-training, recovers source-checkpoint quality. We apply the recipe to a continued pre-trained checkpoint of LFM2-8B-A1B, an 8B-parameter Mixture-of-Experts model, to help produce LFM2.5-8B-A1B with a 128K tokenizer. The expanded tokenizer encodes Hindi and Vietnamese in roughly $2.4\times$ and $2.6\times$ fewer tokens than the source (up to $4.0\times$ on Thai). Combining these reductions with the measured per-token cost of the larger vocabulary, we estimate a $2.2$--$3.7\times$ per-character decode speedup for these languages across our reference devices. We release the model weights and the expanded tokenizer, and report the negative findings that shaped the recipe.

\vspace{0.7\baselineskip}
{\footnotesize
\setlength{\tabcolsep}{3pt}%
\begin{tabular}{@{}l l@{}}
\textbf{Publication Date:} & \reportdate \\
\textbf{Correspondence:}   & \correspondingemail
\end{tabular}
}
\end{abstractbox}

\vspace{1.2\baselineskip}

\begin{figure}[h]
    \centering
    \begin{tikzpicture}[
        node distance=0.6cm and 0.5cm,
        stage/.style={
            rectangle, rounded corners=3pt,
            minimum width=2.2cm, minimum height=1.2cm,
            align=center, font=\footnotesize\bfseries,
            draw=liquidpurple, fill=white, line width=0.6pt
        },
        ministage/.style={
            rectangle, rounded corners=3pt,
            minimum width=2.0cm, minimum height=0.95cm,
            align=center, font=\scriptsize\bfseries,
            draw=liquidpurple, fill=white, line width=0.6pt
        },
        model/.style={
            ellipse, minimum width=1.6cm, minimum height=0.9cm,
            align=center, font=\scriptsize,
            draw=black, fill=white, line width=0.6pt
        },
        smallmodel/.style={
            ellipse, minimum width=1.5cm, minimum height=0.75cm,
            align=center, font=\scriptsize,
            draw=black, fill=white, line width=0.5pt
        },
        data/.style={
            rectangle, rounded corners=2pt,
            minimum width=1.4cm, minimum height=0.6cm,
            align=center, font=\scriptsize,
            draw=black, fill=white, line width=0.4pt
        },
        arrow/.style={-{Stealth[length=1.8mm]}, line width=0.5pt, draw=black},
        sectionlabel/.style={font=\scriptsize\itshape, text=gray}
    ]

    \node[model] (oldmodel) {Pre-trained\\Model};
    \node[ministage, right=of oldmodel, text=liquidpurple] (init) {Embedding\\Init};
    \node[ministage, right=of init, text=liquidpurple] (s1) {Stage 1\\Embedding\\Only};
    \node[ministage, right=of s1, text=liquidpurple] (s2) {Stage 2\\CPT};
    \node[ministage, right=of s2, text=liquidpurple] (s3) {Mid-train\\+ Post};
    \node[model, right=of s3] (final) {Final\\Model};

    \node[smallmodel, above=1.3cm of init, xshift=-2.5cm] (oldtok) {Source\\tokenizer\\(65K)};
    \node[ministage, right=of oldtok, text=liquidpurple] (bpe) {Continued\\BPE merges};
    \node[smallmodel, right=of bpe] (newtok) {Expanded\\tokenizer\\(128K)};
    \node[data, above=0.35cm of bpe] (multidata) {Multilingual\\corpus};

    \node[data, below=0.45cm of s1] (d1) {Multilingual\\embedding-tuning};
    \node[data, below=0.45cm of s2] (d2) {Multilingual\\pre-training};
    \node[data, below=0.45cm of s3] (d3) {Mid-train + SFT};

    \node[sectionlabel, left=0.1cm of oldtok] (toplabel)
        {\rotatebox{90}{Tokenizer construction}};
    \node[sectionlabel, left=0.1cm of oldmodel] (botlabel)
        {\rotatebox{90}{Model adaptation}};

    \draw[arrow] (oldtok) -- (bpe);
    \draw[arrow] (bpe) -- (newtok);
    \draw[arrow] (multidata) -- (bpe);

    \draw[arrow, dashed] (newtok.south) -- ++(0,-0.4) -| (init.north);

    \draw[arrow] (oldmodel) -- (init);
    \draw[arrow] (init) -- (s1);
    \draw[arrow] (s1) -- (s2);
    \draw[arrow] (s2) -- (s3);
    \draw[arrow] (s3) -- (final);

    \draw[arrow] (d1) -- (s1);
    \draw[arrow] (d2) -- (s2);
    \draw[arrow] (d3) -- (s3);

    \end{tikzpicture}
    \caption{\textbf{Tokenizer expansion pipeline.} \emph{Top:} the expanded tokenizer is constructed by continuing the source tokenizer's BPE merge procedure on a multilingual corpus, under which most source tokens carry over one-to-one into the expanded vocabulary. \emph{Bottom:} the pre-trained model's embedding matrix is reinitialized for the new vocabulary (rows copied directly for tokens that carry over, mean of sub-token embeddings for new rows), then adapted in two stages (embedding-only training with the rest of the model frozen, then full-model continued pre-training on a balanced multilingual mixture), after which mid-training and post-training proceed to produce the final model. The dashed arrow indicates that the expanded tokenizer is what fixes the new embedding-matrix shape for the entire adaptation pipeline.}
    \label{fig:overview}
\end{figure}
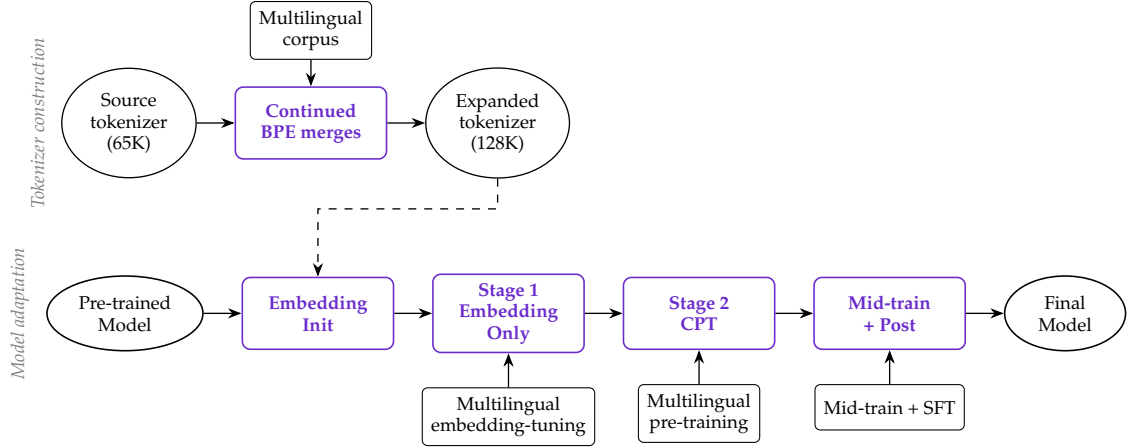

\section{Introduction}
\label{sec:intro}

Tokenizers chosen at the start of pre-training typically allocate vocabulary in proportion to the distribution of the pre-training corpus, which reflects deployment priorities at that time. As those priorities shift, languages that were underrepresented originally, or became relevant only later, are split into many more tokens per word than the languages the tokenizer was trained for~\citep{petrov2023language, rust2021good}. Because generation invokes the decoder once per output token, this fragmentation raises decode latency, compute, and energy consumption for users of those languages. For large cloud models, the embedding and LM-head matrices are a negligible fraction of total parameters, so broader script coverage can be bought with a larger vocabulary. On-device models face tighter constraints: the output LM-head projection touches the full vocabulary dimension on every decode step, and at batch size~1 decode is generally memory-bandwidth-bound~\citep{xu2024device}, so a larger vocabulary noticeably increases both the on-device footprint (the embedding/LM-head matrix that must be held in memory) and the per-token decode cost. Vocabulary size is therefore a first-order design constraint, and edge models typically ship with compact vocabularies covering a chosen set of priority languages and domains, accepting fragmentation outside that set.

Open model families resolve the trade-off between vocabulary size and language coverage at different points, from compact tokenizers to much larger ones \citep{grattafiori2024llama3, abdin2024phi3, gemmateam2025gemma3, yang2025qwen3}. LFM2~\citep{liquid2025lfm2} sits at the compact end, as expected for an on-device model. Its tokenizer is a 65K-vocabulary byte-level BPE tokenizer optimized for English, code, and a fixed set of additional languages. It allocates little of that budget to several languages that it encodes inefficiently, such as Hindi and Vietnamese, which it therefore splits into several times as many tokens as comparable English text. Two standard remedies are unattractive. Training a new tokenizer and pre-training from scratch discards the compute already invested in the checkpoint. Replacing the tokenizer post-hoc with a zero-shot transfer method~\citep{minixhofer2024zett} solves a harder problem than necessary: it must infer the source-to-target relationship for an \emph{arbitrary} target tokenizer.

We present an approach (Figure~\ref{fig:overview}) for in-place tokenizer expansion, targeting a narrower but common production setting in which the model producer controls the tokenizer upgrade. We initialize the merge table from the pre-existing tokenizer, hold the existing merges fixed, and append new merges learned from a multilingual corpus. This continued-BPE construction maps most source tokens one-to-one into the expanded vocabulary and gives every new token an exact decomposition into source tokens, neither of which is available to methods that target an independently trained third-party tokenizer. The model is initialized to match: rows for the carried-over tokens are copied unchanged, and new rows take the mean of their source-token decomposition. We then adapt the model in two stages: embedding-only training with the body frozen, followed by full-model continued pre-training on a balanced multilingual corpus. We refer to this procedure as \emph{tokenizer expansion}, a form of cross-lingual vocabulary adaptation (CVA)~\citep{yamaguchi2024cva, mundra2024empirical} in which the tokenizer's BPE merges are extended rather than vocabulary entries merely appended. 

We apply the recipe during continued pre-training of an LFM2-8B-A1B checkpoint. The adapted model then goes through the standard mid-training and post-training phases to produce \textbf{LFM2.5-8B-A1B}~\citep{liquid2025lfm2, liquid2026lfm25blog}. The token-count reduction is large: the expanded tokenizer encodes Hindi in roughly $2.4\times$ fewer tokens and Vietnamese in $2.6\times$ fewer than the source tokenizer for the same text, with the largest reduction on Thai at $4.0\times$ (Figure~\ref{fig:fertility}). Combined with the measured per-token cost of the larger vocabulary, we estimate this compression yields a $2.2$--$3.7\times$ per-character decode speedup on these languages across our reference devices (Section~\ref{sec:latency}). 

Our contribution is on the model and deployment side. To our knowledge, ours is the first account of the model-side recipe for in-place, continued-BPE tokenizer expansion at production scale (an 8B-parameter on-device MoE). We present an embedding-initialization and two-stage adaptation recipe that applies the vocabulary upgrade to an existing checkpoint without sacrificing accumulated quality (Section~\ref{sec:eval}), an on-device latency characterization that separates tokenizer compression from the per-token cost of a larger vocabulary (Section~\ref{sec:latency}), and the negative findings that shaped the recipe (Section~\ref{sec:ablations}). The continued-BPE construction itself is shared with independent, concurrent work~\citep{purason2025teaching}, which studies its tokenizer-side properties. We build the model-side recipe on top of it and release the model weights and the expanded tokenizer.

\section{Tokenizer Construction}
\label{sec:tokenizer}
 
The expanded tokenizer is a byte-level BPE tokenizer \citep{sennrich2016neural, radford2019language} built by continuing the merge training of the LFM2 tokenizer on a multilingual corpus weighted toward the under-tokenized languages.
 
\paragraph{Source tokenizer.} The LFM2 tokenizer has a 65K-token byte-level BPE vocabulary trained on a corpus weighted toward English, code, and a fixed set of additional languages (Arabic, Chinese, French, German, Japanese, Korean, Spanish) \citep{liquid2025lfm2}. Its vocabulary predates the broader multilingual deployment requirements that motivate this work.
 
\paragraph{Extended training and mapping properties.} We build the expanded 128K tokenizer by initializing the merge table from the LFM2 tokenizer and continuing BPE training on a multilingual corpus that adds coverage of under-represented languages such as Hindi and Thai, while retaining English and code to preserve compression on existing workloads. Existing merges are kept and new merges are appended, so every new token decomposes, by construction, into a sequence of one or more source tokens. Although the expanded vocabulary reassigns token IDs, 63{,}151 of the 64{,}400 source tokens still appear as single tokens in it (Table~\ref{tab:mapping_summary}), so their embedding rows transfer directly. Every other expanded token splits into two or more source tokens. Section~\ref{sec:embed_init} uses this decomposition for embedding initialization. Further construction details, including special-token handling and a comparison against other production tokenizers, are given in Appendix~\ref{app:tokenizer_construction}.
 
\begin{table}[h]
\centering
\small
\begin{tabular}{lr}
\toprule
\textbf{Metric} & \textbf{Value} \\
\midrule
Expanded tokenizer matrix size (nominal)  & 128{,}000 \\
Expanded tokenizer active entries        & 125{,}017 \\
Source tokenizer matrix size (nominal)   & 65{,}536 \\
Source tokenizer active entries          & 64{,}400 \\
Tokens with 1:1 mapping                  & 63{,}151 \\
Tokens with 1:$N$ mapping ($N \geq 2$)   & 61{,}866 \\
Maximum decomposition length             & 64 \\
\bottomrule
\end{tabular}
\caption{\textbf{Mapping properties between source and expanded tokenizers.} We use the nominal sizes (65K source, 128K expanded) as round numbers throughout the report and the active-entry counts (64{,}400 and 125{,}017) when precision matters. Roughly half of the expanded vocabulary corresponds one-to-one to a single source token. Each of these tokens has a different integer ID in the two vocabularies but the same surface string, so its embedding row is copied from the matching source row. The remaining tokens decompose into sequences of two or more source tokens, of length up to 64, and their embeddings are initialized by averaging (Section~\ref{sec:embed_init}). The remaining 1{,}249 source entries (64{,}400 minus 63{,}151) are reserved or special-token slots that do not carry over as ordinary tokens. This does not affect embedding initialization.}
\label{tab:mapping_summary}
\end{table}
 
\paragraph{Why continued-BPE extension rather than vocabulary union.} An alternative, used by early language-specific Llama derivatives such as Chinese-LLaMA \citep{cui2023chinesellama}, trains a separate tokenizer on the target language and takes the union of its vocabulary with the source vocabulary. Because the two tokenizers are trained independently, a new token need not decompose into source tokens, so there is no canonical source-side decomposition from which to initialize its embedding. Continued-BPE extension keeps a single merge table, so every new token decomposes deterministically into source tokens, which is exactly what the embedding initialization in Section~\ref{sec:embed_init} requires.
 
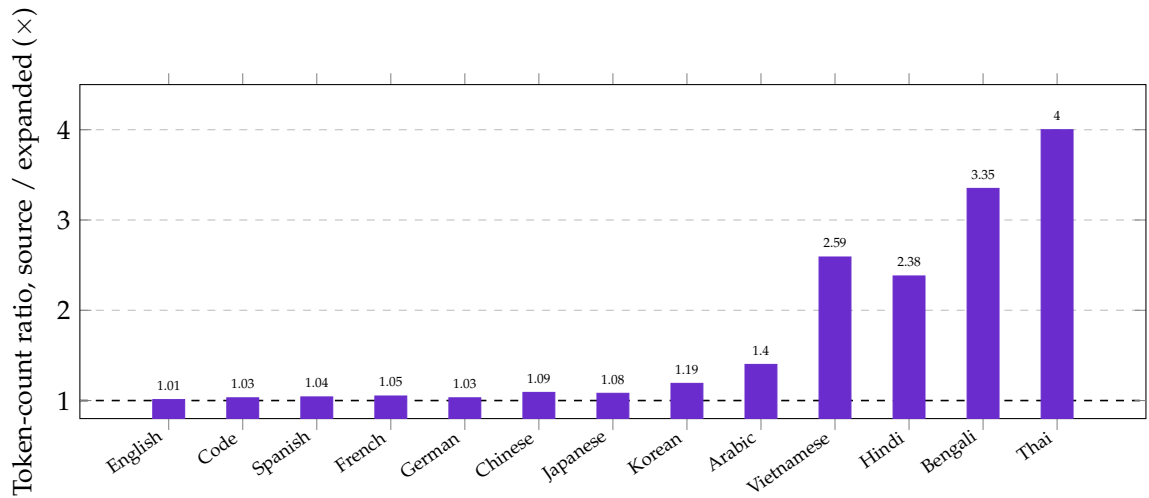
\begin{figure}[h]
    \centering
    \begin{tikzpicture}
    \begin{axis}[
        ybar,
        bar width=12pt,
        width=0.95\textwidth,
        height=6cm,
        ymin=0.8,
        ymax=4.5,
        ylabel={Token-count ratio, source / expanded ($\times$)},
        symbolic x coords={English,Code,Spanish,French,German,Chinese,Japanese,Korean,Arabic,Vietnamese,Hindi,Bengali,Thai},
        xtick=data,
        x tick label style={rotate=35, anchor=east, font=\scriptsize},
        ytick={1,2,3,4},
        extra y ticks={1.0},
        extra y tick labels={},
        extra y tick style={grid=major, grid style={dashed, black, line width=0.6pt}},
        legend pos=north west,
        legend style={font=\scriptsize},
        nodes near coords,
        nodes near coords align={vertical},
        every node near coord/.append style={font=\tiny},
        ymajorgrids=true,
        grid style=dashed,
    ]
    \addplot[fill=liquidpurple, draw=liquidpurple] coordinates {
        (English,1.01) (Code,1.03) (Spanish,1.04) (French,1.05) (German,1.03)
        (Chinese,1.09) (Japanese,1.08) (Korean,1.19) (Arabic,1.40)
        (Vietnamese,2.59) (Hindi,2.38) (Bengali,3.35) (Thai,4.00)
    };
    \end{axis}
    \end{tikzpicture}
    \caption{\textbf{Encoding-length reduction of the expanded tokenizer over the source.} For each language, the bar is the ratio of source-tokenizer to expanded-tokenizer token count on the same input text. A value of 1.0$\times$ is parity and higher means a shorter encoding. English and code are at parity by design, previously supported languages improve modestly, and the under-tokenized languages improve substantially.}
    \label{fig:fertility}
\end{figure}

\paragraph{Language coverage and fertility.} We prioritized languages by how poorly the source tokenizer encodes them. Token fertility is the average number of tokens a tokenizer uses per word, and the source tokenizer has high fertility precisely on the languages that received little vocabulary budget, splitting their words into many tokens. The under-tokenized languages span non-Latin scripts (for example Hindi, Thai, and Bengali) and Latin-script languages poorly served by the source tokenizer. Figure~\ref{fig:fertility} reports, per language, the ratio of source-tokenizer to expanded-tokenizer token count on the same held-out text (non-English web text drawn from FineWeb~2 \citep{penedo2025fineweb2}, plus held-out English, code, and JSON samples), so 1.0$\times$ is parity and higher means a shorter encoding. The under-tokenized languages improve substantially (for example Hindi $2.4\times$ and Thai $4.0\times$), while English and code stay at parity by design. Because batch-1 decode cost is dominated by per-token weight reads, this compression largely carries over into decode-time speedup, as discussed in Section~\ref{sec:latency}.
 
\section{Embedding Initialization}
\label{sec:embed_init}

Given the pre-trained input embedding matrix $E_{\text{src}} \in \mathbb{R}^{|V_{\text{src}}| \times d}$ for the 65K source vocabulary, we build the embedding matrix $E_{\text{new}} \in \mathbb{R}^{|V_{\text{new}}| \times d}$ for the 128K expanded vocabulary ($|V_{\text{new}}| = 125{,}017$ active entries, Table~\ref{tab:mapping_summary}) with a single rule. Each new token $t$ decomposes into a sequence $(t'_1, \ldots, t'_N)$ of source tokens (Section~\ref{sec:tokenizer}), and we set its row to the mean of the corresponding source rows:
\begin{equation}
E_{\text{new}}[t] = \frac{1}{N} \sum_{i=1}^{N} E_{\text{src}}[t'_i].
\end{equation}
For the 63{,}151 tokens that decompose to a single source token ($N=1$) this copies the source row unchanged, and for the rest it averages two or more rows.

Because LFM2-8B-A1B and LFM2.5-8B-A1B tie the input embedding and the output LM head, the same matrix $E_{\text{new}}$ provides both, so initializing a row sets both the token's input representation and its output classifier vector. The same construction applies to untied models, which initialize the two matrices separately.

\paragraph{Why the mean.} We average rather than sum so that new rows keep a norm comparable to existing rows rather than growing with $N$. Because every new token decomposes into known source tokens, this centroid is always defined and no row needs random initialization. We do not claim the centroid is the model's internal representation of the token, only that it is a sensible starting point from which Stage~1 training converges cleanly (Section~\ref{subsec:eval_trajectory}). Initializing new rows by averaging existing embeddings is an established baseline \citep{hewitt2021init, mundra2024empirical}. More elaborate weighted or learned initializers were developed mainly for tokenizer replacement, where the new tokenizer is trained independently and new tokens have no canonical decomposition into source tokens (Section~\ref{sec:related_work}). Here every new token has such a decomposition and Stage~1 trains the new rows, so we did not expect a more elaborate initializer to change the outcome and did not pursue it.
\section{Training Recipe}
\label{sec:training}

The source model for the expansion pipeline is an internal continuation of LFM2-8B-A1B (8.3B total parameters, 1.5B active per token), continued pre-trained from its released 12T-token base \citep{liquid2025lfm2} to 35T tokens, still using the original 65K-vocabulary tokenizer. We call this checkpoint the \emph{source model} throughout. Our primary evidence that the recipe preserves pre-expansion quality is the comparison of the expanded model to this source model.

After embedding initialization, the model is adapted to the expanded tokenizer in two stages, illustrated in Figure~\ref{fig:overview} (lower row of the pipeline) and summarized in Table~\ref{tab:training_summary}, after which mid-training and post-training follow. The staged structure is empirical: training all parameters at once degraded the parts of the model that already worked. Restricting the first stage to the new embedding rows isolates the change, and the subsequent full-model continued pre-training re-balances the model on data tokenized with the expanded tokenizer. As an MoE model, it requires healthy expert routing. We monitored expert load balancing and router health throughout the adaptation, with the router frozen in Stage 1 and trainable in Stage 2 and mid-training. These internal diagnostics, not reported here, showed no degradation from the vocabulary expansion.

\subsection{Stage 1: Embedding-only Training}
\label{subsec:stage1}

In Stage 1 we train the new rows of the tied embedding/LM-head matrix $E_{\text{new}}$ over 600B tokens while keeping all other parameters frozen, including the rows copied directly from the source (the $N=1$ case of Section~\ref{sec:embed_init}). Because the input embeddings and output head are tied, updating a new row changes both the input representation for that token and its output-logit vector. The training data is a multilingual mixture weighted toward the under-tokenized languages whose new rows are being trained, with the standard next-token prediction objective. These 600B tokens train only the embedding and we exclude them from the model's pre-training-token total. We train at a 4{,}096-token sequence length, matching the source model's pre-training context length. Context-length extension to 32K and then 128K is deferred to mid-training (Section~\ref{subsec:stage3}). We reinitialize the optimizer state, warm up the learning rate to a relatively high peak, and then hold it constant. The high peak suits training a small set of freshly initialized parameters.

This embedding-only step is closest to ReTok \citep{gu2024retok}, which also trains only the embedding after a vocabulary change (full comparison in Section~\ref{sec:related_work}). We differ by additionally freezing the copied rows: they already reproduce the source model's lookup and logits for those tokens, and we find they drift harmfully when the rest of the body cannot follow (Section~\ref{sec:ablations}).

Stage 1 alone recovers most of the dip from the zero-shot vocabulary swap. On the per-stage trajectory in Figure~\ref{fig:recovery}, it closes 4.8 of the 5.8 aggregate points lost at the swap, with English and code largely recovered. The under-tokenized languages also recover from the zero-shot dip, but their largest gains come in Stage 2. Stage 1 alone is not sufficient: the full-model continued pre-training of Stage~2 is needed to close the remaining gap and integrate the new vocabulary into the body of the model (Section~\ref{sec:eval}).

\subsection{Stage 2: Continued Pre-training}
\label{subsec:stage2}

In Stage 2 we unfreeze all parameters and continue pre-training for 400B tokens, at the same sequence length as Stage 1, on a balanced multilingual corpus tokenized with the expanded tokenizer. As in Stage 1, we reinitialize the optimizer state and warm up the learning rate before holding it constant, here at a lower peak because the full model starts near a working configuration.

The Stage 2 data mixture is the main factor in whether the unfrozen training preserves what Stage 1 recovered. We studied its effect directly as an ablation (Section~\ref{sec:ablations}), where a predominantly English mixture left multiple-choice accuracy intact but caused a broad regression on generative benchmarks. A mixture with substantially more multilingual content, including coverage of the under-tokenized languages, removed the regression and yielded a checkpoint that matched or exceeded the source model on aggregate benchmarks (Section~\ref{sec:eval}).

\begin{table}[h]
\centering
\small
\begin{tabular}{lrrl}
\toprule
\textbf{Stage} & \textbf{Trainable params} & \textbf{Tokens} & \textbf{Data mixture} \\
\midrule
Stage 1: Embedding only & New rows of $E_{\text{new}}$ & 600B & Under-tokenized focus \\
Stage 2: CPT            & All parameters               & 400B & Balanced multilingual \\
Mid-training             & All parameters               & 2.4T & 32K then 128K context, higher-quality \\
\bottomrule
\end{tabular}
\caption{\textbf{Training summary across the two adaptation stages and the subsequent mid-training.} The 600B Stage~1 tokens train only the embedding and are excluded from the model's pre-training total. The Stage~1 mixture is weighted toward the under-tokenized languages whose new rows are being trained, while the Stage~2 mixture is balanced across the full language set so that full-model training does not regress the previously supported languages (Section~\ref{subsec:stage2}). Mid-training runs in two context-length phases, 32K over 2T tokens then 128K over 400B tokens. Post-training follows the LFM2.5 pipeline \citep{liquid2026lfm25blog}.}
\label{tab:training_summary}
\end{table}

\subsection{Mid-training and Post-training}
\label{subsec:stage3}

Mid-training and post-training follow the LFM2.5 pipeline \citep{liquid2026lfm25blog}, an update of the earlier LFM2 procedure \citep{liquid2025lfm2}. Tokenizer expansion requires no change to this pipeline: the Stage~2 checkpoint is a drop-in replacement for the base model the pipeline normally consumes. Mid-training proceeds in two context-length phases, a 32K-context phase over 2T tokens followed by a 128K-context extension over 400B tokens, both drawing on higher-quality and naturally long-context sources, with the learning rate annealed. The chat template, special tokens, and tool-use formats follow the released LFM2.5 configuration.

\section{Related Work}
\label{sec:related_work}

Tokenizer expansion sits between \emph{cross-lingual vocabulary adaptation} (CVA), which adds language coverage to a pre-trained model by expanding the vocabulary and continuing pre-training, and \emph{tokenizer transfer}, which replaces a model's tokenizer outright and retrains or transfers its embeddings. Prior work establishes the deployment cost of poor tokenization \citep{petrov2023language, ahia2023languages} and studies CVA's inference efficiency and embedding-initialization choices \citep{yamaguchi2024cva, mundra2024empirical}. We position our recipe against the most directly comparable works by the component each one shares with it.

\paragraph{Tokenizer construction.}
The continued-BPE construction we use, initializing the merge table from the source tokenizer and continuing BPE merge training on the target corpus, is also introduced in independent, concurrent work by \citet{purason2025teaching}, who study its tokenizer-side properties (compression, avoidance of unreachable tokens, leaf-based pruning) across several base models at smaller scale and release a toolkit. Our contribution is on the model side: we apply the construction at 8B MoE scale and develop the embedding initialization and two-stage adaptation that recover quality, together with the on-device latency analysis and the accompanying negative findings. \citet{dagan2024getting} show that swapping or substantially altering a pre-trained model's tokenizer requires extensive continued pre-training to avoid performance degradation, which supports the full-model continued pre-training in our Stage~2.

\paragraph{Adaptation recipe.}
Two model-side works are each closest to a single stage of our recipe. \textbf{ReTok} \citep{gu2024retok} trains only the embedding after a tokenizer change with the body frozen, as in our Stage~1, but it replaces the tokenizer and initializes new rows by heuristic segmentation under the old tokenizer. Our continued-BPE construction makes that decomposition exact, and we additionally unfreeze the body in Stage~2, which we find necessary at scale (Section~\ref{sec:ablations}). \textbf{EEVE} \citep{kim2024eeve} progressively unfreezes the model across seven stages, similar to our staged structure, but appends merges from a separately trained tokenizer rather than continuing the source merges.

\paragraph{Embedding initialization.}
The mean-of-subwords centroid we use is a simple, established baseline \citep{hewitt2021init, mundra2024empirical}, and it suffices here because the continued-BPE construction makes the source-side decomposition exact. A large body of work develops more elaborate initializers, using alignment, overlap statistics, weighted source combinations, hypernetworks, or distillation, for the harder setting where the new tokenizer is trained independently of the source and the source-target relationship must be recovered \citep{minixhofer2022wechsel, dobler2023focus, ostendorff2023clptransfer, remy2024transtokenizer, sharthak2025tokenadapt, li2025tokalign, nakash2025adaptivocab, minixhofer2024zett, goddard2025omp, dobler2025tokendistillation}.

\paragraph{Vocabulary extension with continued pre-training.}
Vocabulary extension followed by continued pre-training is well established for language-specific Llama derivatives \citep{cui2023chinesellama, fujii2024swallow} and for multilingual continued training of Southeast Asian \citep{dou2024sailor, dou2025sailor2}, low-resource \citep{yong2023bloom1, lin2024mala500}, instruction-tuned \citep{ustun2024aya}, and translation-oriented \citep{alves2024tower} models. \citet{yamaguchi2025lowresource} show competitive target-language quality from as little as $\sim$0.01\,GB of continued pre-training text. We target a less-explored point in this space: an in-place upgrade of a single already-deployed on-device model, where the central challenge is preserving the model's existing quality while improving its tokenization.

\section{Evaluation}
\label{sec:eval}

We evaluate along two axes. First, quality recovery across the expansion stages (Section~\ref{subsec:eval_trajectory}) tracks an aggregate of pre-training-style benchmarks from the source through the zero-shot swap to Stages~1 and~2. Because these benchmarks are dominated by English, code, and high-resource languages, this axis is primarily evidence that expansion preserves the source checkpoint's existing capability, rather than a measure of the newly added languages. Second, per-language detail on Global-MMLU (Section~\ref{subsec:eval_multilingual}) covers both sides of the goal: it confirms that previously supported languages do not regress and quantifies the gains on the under-tokenized languages that motivated the expansion. Beyond these two axes, the released LFM2.5-8B-A1B also improves over LFM2-8B-A1B on instruction following, tool use, math, and agentic tasks. Those gains reflect the full pipeline of additional pre-training and reasoning-oriented post-training rather than tokenizer expansion alone, and are reported in the LFM2.5 release materials \citep{liquid2026lfm25blog} rather than here.

\subsection{Quality Recovery Across Stages}
\label{subsec:eval_trajectory}

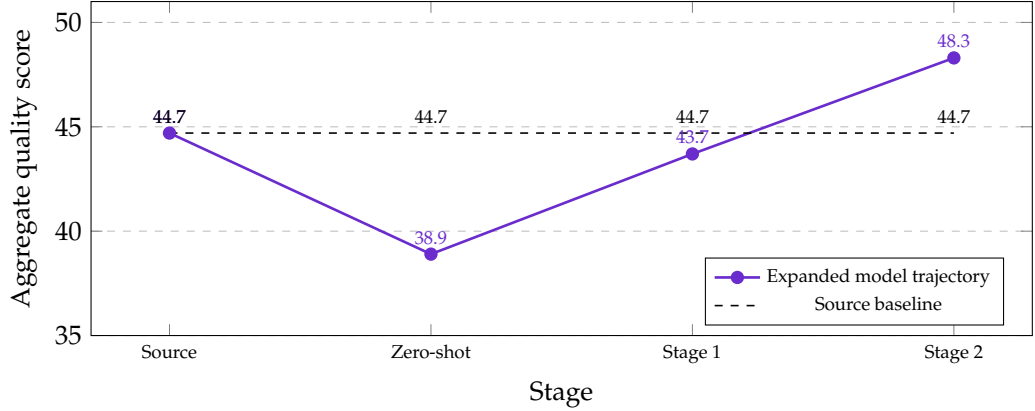
\begin{figure}[h]
    \centering
    \begin{tikzpicture}
    \begin{axis}[
        width=0.85\textwidth,
        height=6cm,
        xlabel={Stage},
        ylabel={Aggregate quality score},
        symbolic x coords={Source,Zero-shot,Stage 1,Stage 2},
        xtick=data,
        x tick label style={font=\scriptsize},
        ymin=35, ymax=51,
        ymajorgrids=true,
        grid style=dashed,
        legend pos=south east,
        legend style={font=\scriptsize},
        nodes near coords,
        nodes near coords align={vertical},
        every node near coord/.append style={font=\scriptsize},
    ]
    \addplot[
        mark=*,
        color=liquidpurple,
        line width=1pt,
    ] coordinates {
        (Source,44.7) (Zero-shot,38.9) (Stage 1,43.7) (Stage 2,48.3)
    };
    \addlegendentry{Expanded model trajectory}

    \addplot[
        dashed,
        color=black,
        line width=0.6pt,
    ] coordinates {
        (Source,44.7) (Zero-shot,44.7) (Stage 1,44.7) (Stage 2,44.7)
    };
    \addlegendentry{Source baseline}
    \end{axis}
    \end{tikzpicture}
    \caption{\textbf{Quality recovery across the adaptation stages.} Aggregate score defined in Section~\ref{subsec:eval_trajectory} (the unweighted mean of eight benchmarks, with per-benchmark detail in Appendix~\ref{app:evaluation}, Table~\ref{tab:recovery_detail}). The zero-shot checkpoint loses about six aggregate points, Stage 1 recovers most of the gap, and Stage 2 closes the remainder and surpasses the source.}
    \label{fig:recovery}
\end{figure}

Figure~\ref{fig:recovery} tracks an aggregate quality score, the unweighted mean of eight public benchmarks (Appendix~\ref{app:evaluation}), across four controlled stages: Source $\to$ Zero-shot $\to$ Stage~1 $\to$ Stage~2. These share one training objective and differ only in the vocabulary change and its recovery training, so the trajectory isolates the tokenizer-expansion pipeline. Mid-training and post-training are excluded, since they would confound the tokenizer change with later capability gains.

Because the source is a base pre-training checkpoint with no instruction tuning, we restrict the aggregate to benchmarks a base model can be scored on without instruction following. These are log-probability benchmarks, which rank fixed answer options by the probability the model assigns them, answer-extraction benchmarks, which read a short final answer out of the model's free-form output, and execution-based code benchmarks. We exclude chat-dependent instruction-following benchmarks (e.g.\ IFEval), which would understate the base checkpoint's quality because it has not been trained for chat-template adherence.

The recovery is monotonic (Figure~\ref{fig:recovery}) after the zero-shot checkpoint. The zero-shot checkpoint loses 5.8 aggregate points, reflecting a body not yet adapted to the new vocabulary. Stage~1 (embedding-only, body frozen) recovers 4.8 of them, and Stage~2 (full continued pre-training) closes the gap and surpasses the source by 3.6 points. Since Stage~2 adds 400B tokens of full-model continued pre-training, we read this surplus as potentially the effect of that additional training rather than of the vocabulary change in isolation. The headline result is therefore preservation: the Stage~2 checkpoint matches or exceeds the source on pre-training-style benchmarks while now using the 128K tokenizer. The under-tokenized-language gains that motivated the expansion are reported separately in Section~\ref{subsec:eval_multilingual}. By category, Stage~2 exceeds the source on Knowledge ($+4.5$), Math ($+2.8$), Code ($+6.3$), and Multilingual ($+1.6$) (Table~\ref{tab:recovery_summary}), though a few individual benchmarks dip slightly, mainly in math and multilingual generation. The Multilingual category here is MMMLU and MGSM, which cover mostly high-resource languages rather than the under-tokenized targets. These dips may reflect the Stage~2 mixture's rebalancing toward the under-tokenized languages rather than a property of the recipe, but we did not isolate the cause. Per-benchmark detail across stages is in Appendix~\ref{app:evaluation}, Table~\ref{tab:recovery_detail}.

\begin{table}[h]
\centering
\small
\setlength{\tabcolsep}{8pt}
\begin{tabular}{lcccc}
\toprule
\textbf{Category} & \textbf{Source} & \textbf{Zero-shot} & \textbf{Stage 1} & \textbf{Stage 2} \\
\midrule
Knowledge (MMLU-Pro)                    & 30.7 & 28.1 & 30.6 & 35.2 \\
Math (GSM8K, MATH500, GSM-Plus)          & 54.4 & 49.6 & 54.3 & 57.2 \\
Code (HumanEval+, LiveCodeBench v5)     & 30.9 & 20.8 & 27.9 & 37.2 \\
Multilingual (MGSM, MMMLU)              & 50.9 & 46.3 & 50.0 & 52.5 \\
\midrule
\textbf{Aggregate (8-benchmark mean)}   & \textbf{44.7} & \textbf{38.9} & \textbf{43.7} & \textbf{48.3} \\
\bottomrule
\end{tabular}
\caption{\textbf{Per-category quality recovery across expansion stages.} Each category is the unweighted mean of its constituent benchmarks. The aggregate is the unweighted mean of all eight benchmarks, so categories with more benchmarks contribute more to it. Stage~2 matches or exceeds the source in all four categories.}
\label{tab:recovery_summary}
\end{table}

\FloatBarrier
\subsection{Multilingual Benchmark Detail}
\label{subsec:eval_multilingual}

Per-language results confirm the headline pattern: previously supported languages hold their ground while the under-tokenized languages gain, so the expansion adds coverage without regressing what already worked.

\paragraph{Global-MMLU (broad multilingual MCQ).}
Global-MMLU \citep{singh2024globalmmlu} renders the full MMLU test set (about $14{,}000$ four-choice questions per language, chance $25\%$) into 42 languages, using a combination of professional, community-verified, and machine translations. We report the 39 languages on which at least one checkpoint exceeds 30\%. The other 3 stay at the floor throughout (Appendix Table~\ref{tab:globalmmlu_full}). The expanded model recovers and exceeds the source by Stage~2 (mean over these 39 languages: 41.4, 37.0, 40.6, 43.7 across Source, Zero-shot, Stage~1, Stage~2, net $+2.3$). The zero-shot vocabulary swap causes a transient dip (mean 37.0, Arabic falling to 32.9) that Stages~1 and~2 recover. By Stage~2 the previously supported languages are preserved, with per-language changes ranging from $-0.3$ (Spanish) to $+1.8$ (Arabic), at or near the sampling noise, while the under-tokenized languages gain substantially (Vietnamese $+11.6$, Indonesian $+9.1$, Hindi $+7.6$, Malay $+6.8$) (Table~\ref{tab:globalmmlu_main}).

\begin{table}[h]
\centering
\small
\setlength{\tabcolsep}{6pt}
\begin{tabular}{lrrrrr}
\toprule
\textbf{Language} & \textbf{Source} & \textbf{Zero-shot} & \textbf{Stage 1} & \textbf{Stage 2} & $\Delta$ \\
\midrule
\multicolumn{6}{l}{\textit{Previously supported}} \\
Arabic           & 51.4 & 32.9 & 50.1 & 53.2 & +1.8 \\
German           & 56.1 & 53.5 & 55.3 & 57.2 & +1.1 \\
Spanish          & 57.5 & 53.8 & 56.6 & 57.2 & -0.3 \\
French           & 56.5 & 53.1 & 55.6 & 57.4 & +0.9 \\
Japanese         & 53.5 & 48.4 & 52.2 & 55.1 & +1.6 \\
Korean           & 50.7 & 41.4 & 49.8 & 52.4 & +1.7 \\
Chinese          & 53.9 & 46.5 & 52.2 & 54.9 & +1.0 \\
\midrule
\multicolumn{6}{l}{\textit{Under-tokenized}} \\
Hindi            & 32.0 & 26.2 & 33.8 & 39.6 & +7.6 \\
Vietnamese       & 35.8 & 25.6 & 38.8 & 47.4 & +11.6 \\
Indonesian       & 41.9 & 34.0 & 40.2 & 51.0 & +9.1 \\
Malay            & 38.6 & 32.8 & 37.5 & 45.4 & +6.8 \\
\midrule
\textbf{Mean (39 langs)} & \textbf{41.4} & \textbf{37.0} & \textbf{40.6} & \textbf{43.7} & \textbf{+2.3} \\
\bottomrule
\end{tabular}
\caption{\textbf{Global-MMLU accuracy across expansion stages (5-shot).} Four-choice multiple choice scored by answer-letter log-probability with 5-shot prompting, so chance is 25\%. $\Delta$ is Stage~2 minus Source. The mean is over the 39 languages on which at least one checkpoint exceeds 30\%. The 3 at-floor languages are excluded (Appendix Table~\ref{tab:globalmmlu_full}). Per-language scores carry sampling noise on the order of half a point ($n\approx14{,}000$), so we do not read per-language differences of about a point or less as meaningful. Here ``Zero-shot'' is the expansion stage with no recovery training, not the few-shot setting.}
\label{tab:globalmmlu_main}
\end{table}

\paragraph{Under-tokenized languages.}
Among the under-tokenized languages, Global-MMLU covers Hindi, Bengali, Indonesian, Malay, and Vietnamese, all of which improve at Stage~2. Hindi, Indonesian, Malay, and Vietnamese appear in Table~\ref{tab:globalmmlu_main}, and Bengali improves by $+4.8$ (Appendix Table~\ref{tab:globalmmlu_full}). Thai, another under-tokenized language, is not part of Global-MMLU and is not separately evaluated here.

\FloatBarrier

\section{On-device Latency}
\label{sec:latency}

The primary motivation for the tokenizer-expansion effort is on-device decoding efficiency. The user-facing metric is wall-clock time per character of generated text rather than raw tokens per second, so the expansion is a net win on a given language only if its compression gain there outweighs the added per-token cost of the larger vocabulary. In brief, we estimate that on all three reference devices the expanded tokenizer decodes the heavily under-tokenized languages (Thai, Bengali, Vietnamese, Hindi) 2.2$\times$ to 3.7$\times$ faster per character than the source tokenizer, at a per-character cost of up to about 9\% on the least-compressed languages (English and code). The two components behind these figures, per-language compression and per-device decode cost, are measured separately in the rest of this section.

\paragraph{Why we report a synthesis rather than a direct end-to-end measurement.}
Both LFM2-8B-A1B and LFM2.5-8B-A1B are released, so one direct option is to decode the same Hindi text on the same device with each release and compare. We do not treat that comparison as our primary evidence, because it does not isolate the tokenizer: the two releases differ along many axes beyond it, and LFM2.5-8B-A1B is a reasoning model that emits variable-length thinking traces, so the same prompt yields different output lengths and token distributions. Restricting the change to the tokenizer alone, by swapping it into a single post-trained checkpoint, is not feasible without re-running the full adaptation pipeline, which would introduce changes of its own. Either route therefore confounds the tokenizer change with other simultaneous changes.

Instead, we decompose decode time per character of output into two independently measurable components. The decoding speedup of the expanded tokenizer over the source tokenizer for the same model body factors cleanly:
\begin{equation*}
\underbrace{\frac{\text{chars/sec}_{\text{new}}}{\text{chars/sec}_{\text{old}}}}_{\text{decode speedup per character}}
\;=\;
\underbrace{\frac{\text{chars/token}_{\text{new}}}{\text{chars/token}_{\text{old}}}}_{\text{compression (per-language)}}
\;\times\;
\underbrace{\frac{\text{decode t/s}_{\text{new}}}{\text{decode t/s}_{\text{old}}}}_{\text{vocabulary-size cost (per-device)}}
\end{equation*}
The compression term is a property of the tokenizers alone, measured on multilingual text without running the model (Figure~\ref{fig:fertility}, Section~\ref{sec:tokenizer}). The vocabulary-size cost term is a property of the model on a given device, measured on the same LFM2-8B-A1B body with only the embedding/LM-head matrix size varied (Section~\ref{subsec:latency_vocab_cost}). The factorization is an identity. What makes the synthesized speedup an estimate is that we measure the vocabulary-size cost once and reuse it as a constant across languages and output lengths. That is a good approximation here: at batch size 1, decode is memory-bandwidth-bound and each step is dominated by reading the model weights and the LM-head matrix, a near-constant number of bytes per token regardless of which tokens are generated. The main thing it leaves out is the KV cache, which grows with context length and is identical for both vocabularies, so on long sequences it takes a larger share of each step and the per-token vocabulary penalty shrinks.

\subsection{Per-token decode cost of vocabulary scaling}
\label{subsec:latency_vocab_cost}

To isolate the vocabulary-size cost from the tokenization compression, we hold the model body fixed and vary only the embedding and LM-head matrix size. We benchmark the same LFM2-8B-A1B weights with that matrix resized from the source tokenizer's 65K entries to the expanded tokenizer's 128K entries, feeding identical token sequences in both configurations so that the measurement isolates the per-token decode cost of the larger matrix from any tokenization-induced change in token count. The benchmark uses \texttt{llama-bench} \citep{gerganov_llama_cpp} with a fixed decode length of 128 tokens and Q4\_0 quantization throughout. Prefill is measured with a 512-token prompt on the M4 Max and a 256-token prompt on the Snapdragon 8 Elite Gen 5 phone (shorter to limit thermal throttling), but our synthesis depends only on the decode throughput.

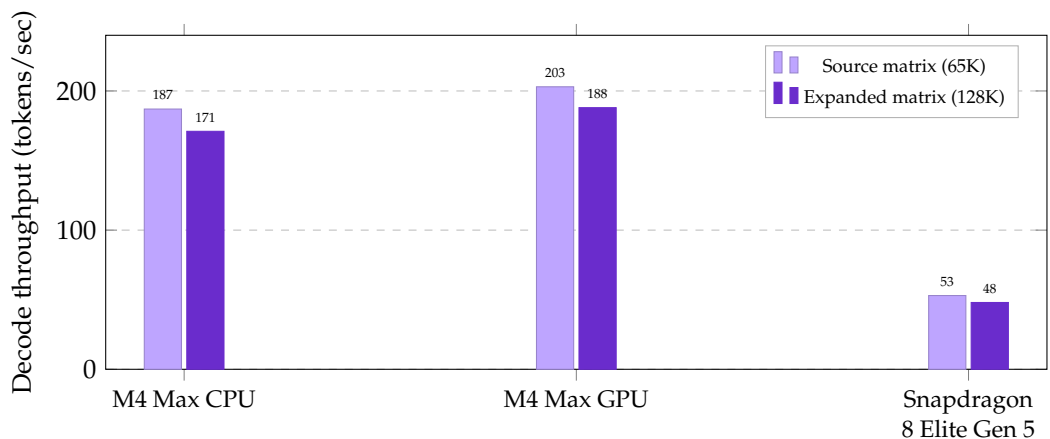
\begin{figure}[h]
    \centering
    \begin{tikzpicture}
    \begin{axis}[
        ybar=2pt,
        bar width=14pt,
        width=0.85\textwidth,
        height=6cm,
        ymin=0,
        ymax=240,
        ylabel={Decode throughput (tokens/sec)},
        symbolic x coords={M4 Max CPU, M4 Max GPU, Snapdragon 8 Elite Gen 5},
        xtick=data,
        xticklabels={M4 Max CPU, M4 Max GPU, {Snapdragon\\8 Elite Gen 5}},
        x tick label style={font=\small, align=center},
        legend pos=north east,
        legend style={font=\scriptsize, draw=gray!60},
        nodes near coords,
        nodes near coords align={vertical},
        every node near coord/.append style={font=\tiny},
        ymajorgrids=true,
        grid style=dashed,
    ]
    \addplot[fill=liquidlavender, draw=liquidlavender!80!black] coordinates {
        (M4 Max CPU,           187)
        (M4 Max GPU,           203)
        (Snapdragon 8 Elite Gen 5,    53)
    };
    \addplot[fill=liquidpurple, draw=liquidpurple] coordinates {
        (M4 Max CPU,           171)
        (M4 Max GPU,           188)
        (Snapdragon 8 Elite Gen 5,    48)
    };
    \legend{Source matrix (65K), Expanded matrix (128K)}
    \end{axis}
    \end{tikzpicture}
    \caption{\textbf{Per-token decode throughput at the source (65K) versus expanded (128K) embedding/LM-head matrix size, identical model body and identical input tokens.} Measured with \texttt{llama-bench} \citep{gerganov_llama_cpp} on LFM2-8B-A1B at Q4\_0 quantization, batch size 1, 128 generated tokens. Decode throughput drops by 8.6\% on the M4 Max CPU, 7.4\% on the M4 Max GPU, and 9.4\% on the Snapdragon 8 Elite Gen 5 phone class when doubling the matrix from 65K to 128K rows. Prefill throughput is within run-to-run variance across matrix sizes, with no consistent direction (full numbers in Table~\ref{tab:vocab_cost}). Model size on disk grows by approximately 100\,MiB with the roughly 62K added embedding rows (65{,}536 to 128{,}000).}
    \label{fig:vocab_cost}
\end{figure}

Prefill and decode respond differently to the larger matrix. Prefill is compute-bound and amortizes the LM-head reads across the input, so it barely moves with vocabulary size. Decode is memory-bandwidth-bound and pays the larger LM-head read on every step, dropping 7--10\% per token. Full per-device numbers and disk sizes are in Appendix~\ref{app:latency_detail}, Table~\ref{tab:vocab_cost}.

\subsection{Per-character decode speedup (synthesis)}
\label{subsec:latency_synthesis}

Combining the per-language compression from Figure~\ref{fig:fertility} with the per-device decode cost ratio from Table~\ref{tab:vocab_cost} gives the estimated decode speedup per character of output, Table~\ref{tab:latency_synthesis}.

\begin{table}[h]
\centering
\small
\setlength{\tabcolsep}{8pt}
\begin{tabular}{lrrrr}
\toprule
\textbf{Language} & \textbf{Compression}    & \multicolumn{3}{c}{\textbf{Synthesized decode speedup per character}} \\
\cmidrule(lr){3-5}
                  & (Fig.~\ref{fig:fertility}) & M4 Max CPU & M4 Max GPU & \shortstack[r]{Snapdragon\\8 Elite Gen 5} \\
\midrule
Thai              & 4.00$\times$ & 3.66$\times$ & 3.70$\times$ & 3.62$\times$ \\
Bengali           & 3.35$\times$ & 3.06$\times$ & 3.10$\times$ & 3.03$\times$ \\
Vietnamese        & 2.59$\times$ & 2.37$\times$ & 2.40$\times$ & 2.35$\times$ \\
Hindi             & 2.38$\times$ & 2.18$\times$ & 2.20$\times$ & 2.16$\times$ \\
Arabic            & 1.40$\times$ & 1.28$\times$ & 1.30$\times$ & 1.27$\times$ \\
Korean            & 1.19$\times$ & 1.09$\times$ & 1.10$\times$ & 1.08$\times$ \\
Chinese           & 1.09$\times$ & 1.00$\times$ & 1.01$\times$ & 0.99$\times$ \\
Japanese          & 1.08$\times$ & 0.99$\times$ & 1.00$\times$ & 0.98$\times$ \\
French            & 1.05$\times$ & 0.96$\times$ & 0.97$\times$ & 0.95$\times$ \\
Spanish           & 1.04$\times$ & 0.95$\times$ & 0.96$\times$ & 0.94$\times$ \\
German            & 1.03$\times$ & 0.94$\times$ & 0.95$\times$ & 0.93$\times$ \\
Code              & 1.03$\times$ & 0.94$\times$ & 0.95$\times$ & 0.93$\times$ \\
English           & 1.01$\times$ & 0.92$\times$ & 0.94$\times$ & 0.91$\times$ \\
\bottomrule
\end{tabular}
\caption{\textbf{Synthesized decode speedup of the expanded tokenizer over the source tokenizer, per character of generated output, by language and device.} Each entry is the product of the per-language compression ratio (Figure~\ref{fig:fertility}) and the per-device decode-throughput ratio (decode t/s of the 128K matrix divided by decode t/s of the 65K matrix, Table~\ref{tab:vocab_cost}). Values above 1.0$\times$ indicate the expanded tokenizer produces the same character output faster. Values below 1.0$\times$ indicate the same output is produced slightly slower.}
\label{tab:latency_synthesis}
\end{table}

Two patterns stand out in Table~\ref{tab:latency_synthesis}. On Thai, Bengali, Vietnamese, and Hindi, where compression is largest, the expanded tokenizer produces 2.2$\times$ to 3.7$\times$ as many characters per unit time across all three reference devices, the primary benefit. Because the per-token cost of the larger LM-head is the same for every language, the change nets out as a gain wherever compression covers it (those four, plus previously supported languages like Arabic and Korean) and as a regression of up to about 9\% per character where compression is near parity (English, code, the major European languages). This is the deliberate trade-off of the design: large gains on under-served scripts, paid for by a small uniform per-token cost. We discuss it further in Section~\ref{subsec:limitations}.

\subsection{Vocabulary-size scaling beyond 128K}
\label{subsec:latency_scaling}

\begin{figure}[h]
    \centering
    \begin{tikzpicture}
    \begin{axis}[
        width=0.85\textwidth,
        height=6cm,
        xlabel={Embedding / LM-head vocabulary size},
        ylabel={Decode throughput / 65K decode throughput},
        ymin=0.55,
        ymax=1.05,
        xmin=48,
        xmax=272,
        xtick={64, 128, 192, 256},
        xticklabels={65K, 128K, 192K, 256K},
        ytick={0.6, 0.7, 0.8, 0.9, 1.0},
        legend pos=south west,
        legend style={font=\scriptsize, draw=gray!60, fill=white, fill opacity=0.9, text opacity=1},
        ymajorgrids=true,
        grid style=dashed,
        line width=1.0pt,
    ]
    \draw[gray!50, dashed] (axis cs:48,1.0) -- (axis cs:272,1.0);
    \addplot[color=liquidlavender!80!black, mark=triangle*, mark size=3pt] coordinates {
        (64, 1.00) (128, 0.93) (192, 0.86) (256, 0.83)
    };
    \addlegendentry{M4 Max GPU}
    \addplot[color=liquidpurple!60!black, mark=square*, mark size=2.5pt] coordinates {
        (64, 1.00) (128, 0.91) (192, 0.85) (256, 0.79)
    };
    \addlegendentry{M4 Max CPU}
    \addplot[color=liquidpurple, mark=*, mark size=2.5pt] coordinates {
        (64, 1.00) (128, 0.91) (192, 0.74) (256, 0.63)
    };
    \addlegendentry{Snapdragon 8 Elite Gen 5}
    \end{axis}
    \end{tikzpicture}
    \caption{\textbf{Matrix-size scaling of decode throughput, 65K to 256K, on identical LFM2-8B-A1B body weights.} Each curve plots the device's decode throughput at the given embedding/LM-head matrix size divided by its 65K decode throughput. The dashed horizontal line marks 65K parity. Decode cost grows with matrix size on every device, steepest on the Snapdragon 8 Elite Gen 5 phone class, where the LM-head projection is the largest fraction of total decode bandwidth. At 256K, the Snapdragon 8 Elite Gen 5 loses 37\% of its 65K decode throughput.}
    \label{fig:vocab_scaling_full}
\end{figure}
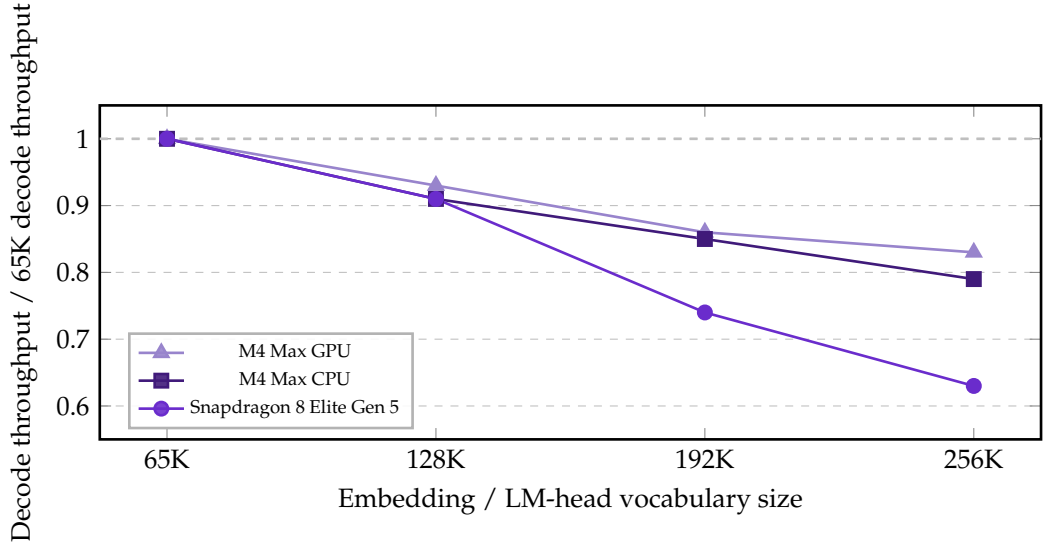

We considered, but did not adopt, larger vocabulary sizes. Figure~\ref{fig:vocab_scaling_full} extends the experiment in Figure~\ref{fig:vocab_cost} to 192K and 256K matrix sizes on the same three devices, plotted as decode-throughput ratio relative to the 65K baseline on each device.

Decode cost rises with vocabulary size on every device, fastest on the Snapdragon 8 Elite Gen 5, where doubling to 128K costs about 9\% of decode throughput, tripling to 192K costs 26\%, and quadrupling to 256K costs 37\%. A larger vocabulary is worth it on a language only if its added compression outweighs this added device cost. Take 256K on the Snapdragon: the decode-cost ratio is 0.63, so a language needs compression above $1/0.63 \approx 1.59\times$ over the source just to break even. Only Thai, Bengali, Vietnamese, and Hindi reach that, and they already clear it at 128K, where the device cost is far lower. The incremental compression from doubling the vocabulary again (128K to 256K) would have to be large to overcome the further throughput loss, and BPE compression has diminishing returns as the vocabulary grows. We therefore stop at 128K on these devices.

\FloatBarrier

\section{Ablations and Negative Findings}
\label{sec:ablations}

Two findings shaped the final recipe. We report them here both for transparency about what did not work and as practical guidance for practitioners attempting similar tokenizer-expansion efforts.

\paragraph{Training the full embedding matrix in Stage 1 degrades performance.}
To decide how much of the embedding to train in Stage 1, we ran the alternative configuration: training all rows of $E_{\text{new}}$, including the rows copied from the source model, rather than only the newly added rows, with the rest of the model frozen in both cases. The hypothesis was that letting the existing rows adjust would help the model reconcile the embedding space with the new vocabulary. Instead it degraded the model, with the damage concentrated on generation: MATH500, HumanEval+, and MGSM all fell while the model began to repeat itself, even though its multiple-choice scores (MMLU and MMMLU) were unchanged or higher. Because the input and output embeddings are tied, training $E_{\text{new}}$ also moves the unembedding that produces the output logits. We did not isolate the mechanism directly, but the most likely explanation is that the copied rows are already matched to the frozen body, and letting them drift while the body cannot adapt breaks that match between the body and the logit layer. The reason the damage shows up in generation while sparing multiple-choice ranking is the same asymmetry we analyze under the Stage~2 mixture ablation below. Restricting Stage 1 to the new rows avoids the problem. Table~\ref{tab:ablation_freeze} reports the comparison.

\begin{table}[h]
\centering
\small
\setlength{\tabcolsep}{6pt}
\begin{tabular}{lrrl}
\toprule
\textbf{Benchmark} & \textbf{New rows only} & \textbf{Full embedding} & \textbf{Reading} \\
                   & \textbf{(final recipe)} & \textbf{matrix}         & \\
\midrule
MMLU                                  & 61.3 & 61.3 & MCQ: knowledge intact \\
MMMLU                                 & 39.5 & 42.6 & MCQ: multilingual reading intact \\
MATH500                               & 49.4 & 27.6 & Generation: collapses \\
HumanEval+                            & 43.9 & 1.8  & Code generation: collapses \\
MGSM                                  & 60.5 & 44.3 & Multilingual generation: declines \\
\midrule
Repetition score ($\uparrow$ better)  & 0.92 & 0.19 & Generation: severe looping \\
\bottomrule
\end{tabular}
\caption{\textbf{Ablation on Stage 1 freezing strategy.} \emph{New rows only} is the final recipe, in which only the newly added embedding rows are trainable (the same Stage~1 checkpoint as Table~\ref{tab:recovery_detail}). \emph{Full embedding matrix} trains all rows of $E_{\text{new}}$, including the rows copied from the source model. The rest of the model is frozen in both cases. Multiple-choice accuracy (MMLU, MMMLU) is unchanged or higher when the full matrix is trained, while every generation benchmark falls and the repetition score drops sharply: the same failure mode discussed under the Stage 2 data-mixture ablation (Table~\ref{tab:ablation_cpt}). The repetition score is an internal diagnostic: the fraction of held-out generations that pass an automated n-gram-loop check (no degenerate repetition), from 0 to 1, where higher is better and lower indicates more looping. The final recipe's $0.92$ is essentially at the source level ($0.93$, Table~\ref{tab:ablation_cpt}).}
\label{tab:ablation_freeze}
\end{table}

\paragraph{Stage 2 data mixture is critical.}
To measure how much the Stage 2 mixture matters, we ran it with a deliberately unbalanced, predominantly English continued pre-training mix, testing the hypothesis that multilingual coverage was already established in Stage 1 and that English data would preserve general capability. The result was not a graceful loss of multilingual quality but a broad generative collapse. The clearest evidence is MATH500, an English-language benchmark, which fell from 61.8 to 27.0. MGSM fell from 61.2 to 28.4, and the model exhibited markedly more degenerate repetition. All of this happened while the model \emph{improved} on log-probability-scored MCQ benchmarks (MMLU and MMMLU). Constructing a more balanced multilingual mixture eliminated the regression and produced the checkpoint reported in Section~\ref{sec:eval}. The comparison is in Table~\ref{tab:ablation_cpt}.

The MCQ-versus-generation asymmetry is the key diagnostic. Multiple-choice benchmarks scored by log-probability over A/B/C/D options measure whether the model assigns higher probability to the correct option than to the distractors. They do not require the model to \emph{generate} coherent text. The bad-mixture model retains, and slightly improves, this discriminative ability, while losing the generative ability needed for free-form math and multilingual reasoning. The MMMLU improvement under the bad mixture fits this story: the model still reads the multilingual prompt correctly and ranks options sensibly, while having lost the ability to produce coherent multilingual output. MMLU-Pro is the partial exception: although multiple-choice in format, it drops with the generative benchmarks (35.2 to 20.6) rather than holding like MMLU and MMMLU.

\begin{table}[h]
\centering
\small
\setlength{\tabcolsep}{6pt}
\begin{tabular}{lrrrl}
\toprule
\textbf{Benchmark} & \textbf{Source} & \textbf{Stage 2 good} & \textbf{Stage 2 bad} & \textbf{Reading} \\
                   &                 & \textbf{mixture}       & \textbf{mixture}      & \\
\midrule
MMLU                              & 63.0 & 64.7 & 64.9 & MCQ: knowledge intact \\
MMMLU                             & 39.3 & 43.8 & 48.3 & MCQ: multilingual reading intact \\
MMLU-Pro                          & 30.7 & 35.2 & 20.6 & MCQ reasoning: degrades \\
MATH500                           & 49.2 & 61.8 & 27.0 & Generation: collapses \\
MGSM                              & 62.5 & 61.2 & 28.4 & Multilingual generation: collapses \\
\midrule
Repetition score ($\uparrow$ better)  & 0.93 & 0.88 & 0.49 & Generation: more looping \\
\bottomrule
\end{tabular}
\caption{\textbf{Ablation on Stage 2 data mixture.} Stage~2 (good mixture) is the run used in the released LFM2.5-8B-A1B. Stage~2 (bad mixture) is an ablation with a predominantly English continued pre-training mix. Multiple-choice benchmarks scored by log-probability (MMLU, MMMLU) are unchanged or even improved under the bad mixture, while generation-style benchmarks (MATH500, MGSM) and an internal repetition score collapse (see the diagnostic discussion above). The repetition score is the same diagnostic as in Table~\ref{tab:ablation_freeze}, where lower values indicate more degenerate output.}
\label{tab:ablation_cpt}
\end{table}

A per-language breakdown of the MGSM collapse, and a tentative script-mediated reading of why Latin-script European languages collapse most sharply, is given in Appendix~\ref{app:ablation_detail}.

Taken together, the two ablations point to a practical lesson about evaluation. Two different interventions, drifting the tied embedding against a frozen body and continued pre-training on an unbalanced mixture, produced the same signature, and a practitioner relying on MMLU-style accuracy alone would have judged both broken checkpoints healthy. Tokenizer expansion, and continued pre-training more generally, can degrade a model in ways that multiple-choice accuracy does not register. This is why the evaluation in Section~\ref{sec:eval} includes generative benchmarks rather than relying on multiple-choice accuracy alone.

\FloatBarrier
\section{Conclusion}
\label{sec:conclusion}

We present tokenizer expansion, an in-place recipe for extending the tokenizer of an already pre-trained LLM in the case where the model producer controls the design of the new tokenizer. By constructing the new tokenizer as a continuation of the source tokenizer's BPE merges (with existing merges held fixed), the embedding initialization problem reduces to direct row copies for the tokens that carry over and the mean of the source sub-token embeddings for the new tokens. A two-stage adaptation (embedding-only training, then unfrozen continued pre-training), followed by the mid-training and post-training phases, preserves source-checkpoint quality on the evaluated suite and substantially reduces token counts on under-tokenized languages. Per-language results show that performance and efficiency on previously supported languages are preserved, while the under-tokenized languages see large tokenizer-efficiency improvements (Section~\ref{sec:latency}), together with multilingual quality gains on the evaluated languages from the Stage~2 data (Section~\ref{subsec:eval_multilingual}).

We used the recipe to help produce LFM2.5-8B-A1B, an updated version of the LFM2-8B-A1B on-device MoE model with a 128K tokenizer. The model is released with open weights. On languages that the source tokenizer fragmented heavily (for example Thai, Hindi, and Vietnamese), the expanded tokenizer needs fewer tokens for the same text, which our synthesis estimates as a substantial per-character decode speedup once the larger vocabulary's per-token cost is accounted for (Section~\ref{sec:latency}). We document two negative findings that shaped the recipe: training the full embedding matrix in Stage 1 degrades performance, and the Stage 2 data mixture must be balanced multilingually to avoid regression at the unfreezing step.

The recipe is narrow by design, suited to the setting where a model producer wishes to extend coverage of an existing checkpoint to additional languages or scripts without restarting training. For that setting, this tokenizer expansion recipe offers a practical reference point alongside the broader literature on cross-lingual vocabulary adaptation and zero-shot tokenizer transfer.

\subsection{Limitations and Scope}
\label{subsec:limitations}

Several constraints bound where the recipe applies and what it costs.

\paragraph{Restricted to continued-BPE constructions.} The principal restriction is that the new tokenizer must be obtained by continuing the source tokenizer's BPE merges on a new corpus (rather than being trained independently). This requires the ability to retrain BPE on a new corpus while keeping the existing merge table fixed, which in turn requires access to the original tokenizer's merge rules and special-token configuration. The recipe does not apply when the new tokenizer is an off-the-shelf tokenizer designed by a third party. For that setting, the zero-shot transfer methods of \citet{minixhofer2024zett, goddard2025omp} are the appropriate tool.

\paragraph{Single model family.} We validate the recipe on one model family (LFM2), the setting in which we control the tokenizer and can apply the upgrade end to end. Beyond the continued-BPE requirement above, the recipe is not specific to this architecture: it assumes only a checkpoint whose embeddings can be extended and trained further. We therefore expect it to transfer to other model families whose producers have the same control over the tokenizer.

\paragraph{Continued pre-training compute.} Stage 2 is a continued pre-training stage and carries a corresponding compute cost. In our runs, this was a small fraction of the original pre-training budget (LFM2-8B-A1B was pre-trained on 12\,T tokens \citep{liquid2025lfm2}, and the LFM2.5 release adds further pre-training to 38\,T total \citep{liquid2026lfm25blog}), but it is not free. We did not determine the minimum continued pre-training budget needed to recover source-checkpoint quality, so this cost reflects our setup rather than a requirement of the method. The recipe is most attractive when the original pre-training cost is large enough that even a fractional re-investment in continued pre-training is preferable to training from scratch.

\paragraph{Decode regression on previously supported languages.} On languages already encoded efficiently by the source tokenizer (English, code, and the major European languages), the larger vocabulary's per-token cost is not offset by a compression gain and shows up as a per-character decode regression of up to about 9\% (Section~\ref{subsec:latency_synthesis}, Table~\ref{tab:latency_synthesis}). The trade-off is favorable for our deployment targets, but practitioners whose workloads are dominated by these languages should weigh it when applying the recipe.

\subsection{Availability}
\label{subsec:avail}
LFM2.5-8B-A1B is released with open weights at \href{https://huggingface.co/LiquidAI/LFM2.5-8B-A1B}{huggingface.co/LiquidAI/LFM2.5-8B-A1B} under the LFM Open License v1.0 along with the expanded 128K tokenizer and deployment guides for llama.cpp, MLX, vLLM, and SGLang.

\section*{Acknowledgments}
We thank the broader Liquid team for the base model and the surrounding infrastructure on which this work depends \citep{liquid2025lfm2}.

\bibliographystyle{plainnat}
\bibliography{references}

\newpage
\appendix

\section{Evaluation Benchmark Details}
\label{app:evaluation}

The per-stage recovery aggregate in Section~\ref{subsec:eval_trajectory} (Figure~\ref{fig:recovery}, Table~\ref{tab:recovery_summary}) is computed over a fixed suite of eight public benchmarks chosen to isolate the tokenizer-expansion contribution rather than downstream mid-training gains. The suite is:

\begin{itemize}
\item \textbf{Knowledge (1 benchmark).} MMLU-Pro \citep{wang2024mmlupro} (we report the answer-ranking variant scored by log-probability over the option set, which is more stable across base and post-trained checkpoints than free-form answer extraction).
\item \textbf{Math (3 benchmarks).} GSM8K \citep{cobbe2021gsm8k}, MATH500 (the 500-problem test subset of MATH \citep{hendrycks2021math} selected by \citet{lightman2023letsverify}), and GSM-Plus \citep{li2024gsmplus}.
\item \textbf{Code (2 benchmarks).} HumanEval+ \citep{liu2023humanevalplus} and LiveCodeBench v5 \citep{jain2024livecodebench}.
\item \textbf{Multilingual (2 benchmarks).} MMMLU \citep{openai2024mmmlu} (the multilingual counterpart of MMLU, scored by log-probability over multiple-choice options, reported on the language subset configured for this evaluation) and MGSM \citep{shi2023language} (multilingual GSM8K, generation-style).
\end{itemize}

We exclude chat-template-dependent benchmarks (IFEval \citep{zhou2023ifeval}, Multi-IF \citep{he2024multiif}, and other tool-use and instruction-following benchmarks) from the Section~\ref{subsec:eval_trajectory} aggregate, for the reason given there: the source pre-training checkpoint has not been trained for chat-template adherence or tool-use formatting, so these benchmarks would understate it. The released LFM2.5-8B-A1B incorporates the corresponding capabilities through post-training (Section~\ref{sec:eval}).

\begin{table}[h]
\centering
\small
\setlength{\tabcolsep}{6pt}
\begin{tabular}{llrrrr}
\toprule
\textbf{Category} & \textbf{Benchmark} & \textbf{Source} & \textbf{Zero-shot} & \textbf{Stage 1} & \textbf{Stage 2} \\
\midrule
Knowledge   & MMLU-Pro              & 30.7 & 28.1 & 30.6 & 35.2 \\
\midrule
Math        & GSM8K                 & 67.7 & 65.1 & 67.1 & 67.0 \\
            & MATH500               & 49.2 & 39.0 & 49.4 & 61.8 \\
            & GSM-Plus               & 46.2 & 44.7 & 46.5 & 42.7 \\
\midrule
Code        & HumanEval+            & 51.8 & 34.2 & 43.9 & 60.4 \\
            & LiveCodeBench v5      & 10.0 &  7.3 & 11.8 & 14.1 \\
\midrule
Multilingual& MMMLU                 & 39.3 & 37.1 & 39.5 & 43.8 \\
            & MGSM                  & 62.5 & 55.4 & 60.5 & 61.2 \\
\midrule
\multicolumn{2}{l}{\textbf{8-benchmark aggregate}} & \textbf{44.7} & \textbf{38.9} & \textbf{43.7} & \textbf{48.3} \\
\bottomrule
\end{tabular}
\caption{\textbf{Per-benchmark recovery across expansion stages.} Detail behind Figure~\ref{fig:recovery} and Table~\ref{tab:recovery_summary}. MMMLU is grouped under Multilingual together with MGSM, matching the main-text categorization in Table~\ref{tab:recovery_summary}. The aggregate is the unweighted mean of all eight per-benchmark values per stage (not the mean of category means). All values are accuracy in percent except LiveCodeBench v5, which is pass@1.}
\label{tab:recovery_detail}
\end{table}

\begin{table}[h]
\centering
\small
\setlength{\tabcolsep}{6pt}
\begin{tabular}{lrrrrr}
\toprule
\textbf{Language} & \textbf{Source} & \textbf{Zero-shot} & \textbf{Stage 1} & \textbf{Stage 2} & $\Delta$ \\
\midrule
Arabic           & 51.4 & 32.9 & 50.1 & 53.2 & +1.8 \\
Bengali          & 27.4 & 25.0 & 28.5 & 32.2 & +4.8 \\
Chinese          & 53.9 & 46.5 & 52.2 & 54.9 & +1.0 \\
Czech            & 41.7 & 40.2 & 41.2 & 43.9 & +2.2 \\
Dutch            & 47.9 & 43.5 & 45.3 & 50.8 & +2.9 \\
English          & 62.3 & 59.8 & 60.4 & 64.1 & +1.8 \\
Filipino         & 37.7 & 34.8 & 36.7 & 40.8 & +3.1 \\
French           & 56.5 & 53.1 & 55.6 & 57.4 & +0.9 \\
German           & 56.1 & 53.5 & 55.3 & 57.2 & +1.1 \\
Greek            & 32.4 & 27.7 & 32.6 & 33.8 & +1.4 \\
Hausa            & 29.9 & 30.0 & 30.3 & 31.1 & +1.2 \\
Hebrew           & 39.7 & 25.7 & 32.3 & 40.7 & +1.0 \\
Hindi            & 32.0 & 26.2 & 33.8 & 39.6 & +7.6 \\
Igbo             & 29.7 & 29.4 & 29.1 & 31.6 & +1.9 \\
Indonesian       & 41.9 & 34.0 & 40.2 & 51.0 & +9.1 \\
Italian          & 57.0 & 52.0 & 55.0 & 57.4 & +0.4 \\
Japanese         & 53.5 & 48.4 & 52.2 & 55.1 & +1.6 \\
Korean           & 50.7 & 41.4 & 49.8 & 52.4 & +1.7 \\
Kyrgyz           & 32.0 & 30.3 & 31.7 & 32.0 & ~0.0 \\
Lithuanian       & 32.5 & 31.5 & 32.2 & 33.5 & +1.0 \\
Malagasy         & 30.4 & 30.2 & 30.2 & 32.2 & +1.8 \\
Malay            & 38.6 & 32.8 & 37.5 & 45.4 & +6.8 \\
Nepali           & 30.4 & 25.8 & 28.8 & 33.5 & +3.1 \\
Nyanja           & 30.5 & 30.2 & 30.4 & 31.2 & +0.7 \\
Persian          & 42.8 & 27.5 & 36.8 & 41.9 & -0.9 \\
Polish           & 49.3 & 45.1 & 49.1 & 51.1 & +1.8 \\
Portuguese       & 56.6 & 51.8 & 55.9 & 56.7 & +0.1 \\
Romanian         & 44.7 & 40.0 & 43.5 & 46.5 & +1.8 \\
Russian          & 51.1 & 47.6 & 50.3 & 52.7 & +1.6 \\
Serbian          & 38.0 & 36.6 & 37.8 & 37.7 & -0.3 \\
Shona            & 31.2 & 30.8 & 31.2 & 32.3 & +1.1 \\
Somali           & 29.5 & 29.0 & 29.6 & 30.8 & +1.3 \\
Spanish          & 57.5 & 53.8 & 56.6 & 57.2 & -0.3 \\
Swahili          & 30.1 & 29.5 & 29.8 & 31.5 & +1.4 \\
Swedish          & 44.5 & 40.8 & 44.6 & 47.4 & +2.9 \\
Turkish          & 36.7 & 31.4 & 36.1 & 42.1 & +5.4 \\
Ukrainian        & 43.0 & 41.7 & 42.6 & 44.1 & +1.1 \\
Vietnamese       & 35.8 & 25.6 & 38.8 & 47.4 & +11.6 \\
Yoruba           & 28.9 & 28.3 & 28.3 & 30.1 & +1.2 \\
\midrule
\textbf{Mean} & \textbf{41.4} & \textbf{37.0} & \textbf{40.6} & \textbf{43.7} & \textbf{+2.3} \\
\bottomrule
\end{tabular}
\caption{\textbf{Global-MMLU accuracy per language across expansion stages (5-shot, answer-letter log-probability, chance $=25\%$).} $\Delta$ is Stage~2 minus Source. Sampling noise is on the order of half a point per language ($n\approx14{,}000$). ``Zero-shot'' denotes the expansion stage with no recovery training, not the few-shot setting. Shown are the 39 languages on which at least one checkpoint exceeds 30\%. Omitted (3, at or near chance for all checkpoints): Amharic, Sinhala, Telugu.}
\label{tab:globalmmlu_full}
\end{table}

\section{Ablation Details}
\label{app:ablation_detail}

This appendix gives the per-language breakdown behind the Stage 2 data-mixture ablation (Section~\ref{sec:ablations}, Table~\ref{tab:ablation_cpt}).

The per-language breakdown of MGSM (Table~\ref{tab:ablation_cpt_mgsm}) sharpens the diagnosis. The drop relative to the good mixture is not uniform across languages. The largest are on Latin-script European languages: French falls from 71.6 to 10.4 ($-$61 points), German from 66.4 to 18.0 ($-$48), and Spanish from 56.4 to 12.4 ($-$44). CJK and Cyrillic languages hold up substantially better: Japanese drops only 7 points, Russian 14, and Chinese 22. We read this as further evidence that the failure mode under the bad mixture is generation behavior rather than lost capability. The observed pattern is consistent with, though does not establish, a script-mediated mechanism: when the model is steered toward English-style output by the predominantly English mixture, languages whose Latin script and shared vocabulary make the boundary with English porous may be more easily shifted toward English-like output, while script-distinct languages (CJK, Cyrillic) retain their language-specific generation behavior more readily. A controlled study with more languages per script group, statistical testing, and a defined linguistic-distance metric would be needed to make this stronger than an observation.

\begin{table}[h]
\centering
\small
\setlength{\tabcolsep}{6pt}
\begin{tabular}{lrrrr}
\toprule
\textbf{Language}     & \textbf{Source} & \textbf{Stage 2 good} & \textbf{Stage 2 bad} & $\Delta$ bad vs good \\
\midrule
Spanish               & 69.6 & 56.4 & 12.4 & $-$44.0 \\
French                & 64.8 & 71.6 & 10.4 & $-$61.2 \\
German                & 60.8 & 66.4 & 18.0 & $-$48.4 \\
Chinese               & 57.2 & 52.0 & 30.0 & $-$22.0 \\
Japanese              & 56.8 & 59.2 & 52.4 & $-$6.8 \\
Russian               & 66.0 & 61.6 & 47.2 & $-$14.4 \\
\midrule
\textbf{Mean}         & \textbf{62.5} & \textbf{61.2} & \textbf{28.4} & \textbf{$-$32.8} \\
\bottomrule
\end{tabular}
\caption{\textbf{Per-language MGSM under the Stage 2 data mixture ablation.} In this six-language MGSM subset, the Latin-script European languages (Spanish, French, German) collapse most sharply ($-$44 to $-$61 points), while CJK and Cyrillic languages hold up better. The pattern is consistent with the bad-mixture model losing the ability to generate text whose script overlaps with English while retaining script-distinct generation more readily.}
\label{tab:ablation_cpt_mgsm}
\end{table}

\section{Tokenizer Construction Implementation Notes}
\label{app:tokenizer_construction}

The expanded tokenizer is constructed by initializing the BPE merge table with the source tokenizer's merges and continuing the merge training on the multilingual corpus. The special-token configuration (chat template special tokens, function-call delimiters, BOS/EOS) is preserved unchanged from the source tokenizer. Byte-level pre-tokenization rules are inherited. The vocabulary is capped at 128K and trimmed to the most frequent merges up to that limit. The target vocabulary holds 125{,}017 active tokens (Table \ref{tab:mapping_summary}), with the remaining slots reserved for future expansion.

When the source tokenizer is implemented in \texttt{tiktoken} or another non-SentencePiece BPE framework, the merge-extension procedure differs in implementation details from the SentencePiece case. We refer the reader to \citet{purason2025teaching} for a careful discussion of the implementation differences between byte-level and SentencePiece-based BPE.

\paragraph{Comparison to other production tokenizers.}
Table~\ref{tab:peer_tokenizers} compares the expanded 128K tokenizer to four widely used production tokenizers on a fixed multilingual evaluation corpus. Two observations frame the rest of the table. First, on the \emph{total} corpus the expanded tokenizer produces the lowest token count of any tokenizer measured (11.3M tokens), despite having a smaller vocabulary than all four (gpt-4o 200K, Mistral Small 3.1 131K, Qwen3 151K, gemma-3 262K). Second, on the under-tokenized languages the expanded tokenizer is competitive with all peers and far better than the source, though it does not displace tokenizers that simply allocate more vocabulary slots to those scripts. Gemma-3 (262K vocabulary, more than 2$\times$ the expanded tokenizer's) wins on Hindi (by roughly 2$\times$), Bengali (by roughly 2.3$\times$), Thai, Indonesian, Vietnamese, and Japanese, an unsurprising consequence of having twice the vocabulary budget to spend. With only 128K entries, the expanded tokenizer still produces the best total compression on a multilingually balanced corpus, wins on Arabic, code, and JSON, and comes close to the best peers on the remaining languages, all while remaining a far smaller embedding matrix to serve on-device.

\begin{table}[h]
\centering
\small
\setlength{\tabcolsep}{4pt}
\begin{tabular}{lrrrrrr}
\toprule
                          & \textbf{Source}     & \textbf{Expanded}     & gpt-4o    & Qwen3    & gemma-3  & Mistral  \\
                          & (65K)               & (128K)               & (200K)    & (151K)   & (262K)   & Small-3.1 (131K)        \\
\midrule
English                   & 953                 & 947                  & 945       & \textbf{938}      & 953      & 956 \\
French                    & 628                 & 599                  & \textbf{588} & 657   & 594      & 591 \\
Spanish                   & 870                 & 838                  & \textbf{777} & 897   & 781      & 819 \\
German                    & 752                 & 728                  & \textbf{686} & 817   & 705      & 732 \\
Chinese                   & 1{,}094             & 1{,}000              & 1{,}017   & \textbf{838} & 914   & 1{,}234 \\
Japanese                  & 708                 & 655                  & 834       & 729      & \textbf{636} & 827 \\
Korean                    & 1{,}247             & 1{,}051              & 1{,}103   & 1{,}241  & 1{,}046  & \textbf{1{,}033} \\
Arabic                    & 3{,}104             & \textbf{2{,}211}     & 2{,}452   & 2{,}757  & 2{,}342  & 2{,}224 \\
\midrule
Vietnamese                & 2{,}958             & 1{,}141              & 1{,}200   & 1{,}170  & \textbf{1{,}120} & 1{,}220 \\
Hindi                     & 3{,}036             & 1{,}278              & 1{,}728   & 2{,}264  & \textbf{651} & 1{,}433 \\
Bengali                   & 4{,}259             & 1{,}272              & 1{,}685   & 2{,}276  & \textbf{541} & 1{,}293 \\
Thai                      & 5{,}708             & 1{,}426              & 2{,}060   & 1{,}792  & \textbf{1{,}145} & 2{,}005 \\
Indonesian                & 1{,}733             & 1{,}204              & 1{,}153   & 1{,}428  & \textbf{1{,}088} & 1{,}258 \\
\midrule
Code (Other)              & 26{,}138            & \textbf{25{,}432}    & 27{,}300  & 38{,}480 & 38{,}747 & 38{,}865 \\
JSON                      & 4{,}775             & \textbf{4{,}227}     & 4{,}842   & 5{,}108  & 5{,}389  & 5{,}157 \\
\midrule
\textbf{Total corpus (M)} & 15.7                & \textbf{11.3}        & 12.0      & 14.8     & 12.7     & 13.9 \\
\bottomrule
\end{tabular}
\caption{\textbf{Tokenizer comparison on a held-out multilingual evaluation corpus (non-English web text from FineWeb~2 \citep{penedo2025fineweb2}, plus held-out English, code, and JSON samples).} Each cell reports the number of tokens produced by the tokenizer on the per-language sub-corpus (lower is better, same text across columns). Bold marks the most efficient tokenizer per row among the six columns shown.}
\label{tab:peer_tokenizers}
\end{table}

\section{On-device Latency Details}
\label{app:latency_detail}

Table~\ref{tab:vocab_cost} reports the full per-device throughput and disk-size numbers behind Figure~\ref{fig:vocab_cost}, comparing the 65K and 128K embedding/LM-head matrix on an otherwise-identical LFM2-8B-A1B body.

\begin{table}[h]
\centering
\small
\setlength{\tabcolsep}{6pt}
\begin{tabular}{lrrrrr}
\toprule
\textbf{Device}            & \textbf{Vocab} & \textbf{Model size (MiB)} & \textbf{Prefill (t/s)} & \textbf{Decode (t/s)} & \textbf{$\Delta$ decode} \\
\midrule
M4 Max CPU                 & 65K   & 4516 & 425  & 187 & n/a       \\
M4 Max CPU                 & 128K  & 4618 & 450  & 171 & $-8.6\%$  \\
\midrule
M4 Max GPU                 & 65K   & 4516 & 2794 & 203 & n/a       \\
M4 Max GPU                 & 128K  & 4618 & 2778 & 188 & $-7.4\%$  \\
\midrule
Snapdragon 8 Elite Gen 5         & 65K   & 4514 & 106  & 53  & n/a       \\
Snapdragon 8 Elite Gen 5         & 128K  & 4617 & 107  & 48  & $-9.4\%$  \\
\bottomrule
\end{tabular}
\caption{\textbf{Throughput of LFM2-8B-A1B with 65K vs 128K embedding/LM-head matrix at Q4\_0.} The model body weights are identical across rows: only the embedding and LM-head matrix size differs. Prefill is measured with a 512-token prompt on the M4 Max and a 256-token prompt on the Snapdragon 8 Elite Gen 5 phone. Decode is measured over 128 generated tokens at batch size 1.}
\label{tab:vocab_cost}
\end{table}

\end{document}